\documentclass[runningheads,english,a4paper]{llncs}
\usepackage{color}
\usepackage{amssymb}
\usepackage{graphicx}

\usepackage{amsmath}
\usepackage{amsmath,bm}

\usepackage{amssymb}
\usepackage[T1]{fontenc}
\usepackage[latin9]{inputenc}
\usepackage{array}
\usepackage{float}
\usepackage{amstext}
\usepackage{graphicx}
\usepackage{float}

\makeatletter

\usepackage[ruled,vlined]{algorithm2e}

\usepackage{enumitem}
\setlist{nolistsep}

\@ifundefined{showcaptionsetup}{}{%
 \PassOptionsToPackage{caption=false}{subfig}}
\usepackage{subfig}

\makeatother

\usepackage{url}
\urldef{\mailsa}\path|Siwar.Jendoubi@yahoo.fr, {Arnaud.Martin,|
\urldef{\mailsb}\path| ludovic.lietard}@univ-rennes1.fr, boutheina.yaghlane@ihec.rnu.tn,|   
\urldef{\mailsc}\path|hend.Benhji@cert.mincom.tn|   
\newcommand{\keywords}[1]{\par\addvspace\baselineskip
\noindent\keywordname\enspace\ignorespaces#1}

\begin{document}

\title{Dynamic time warping distance for message propagation classification
in Twitter%
}
\titlerunning{DTW distance for message propagation classification
in Twitter}

\author{Siwar Jendoubi\inst{1,2,4}, Arnaud Martin\inst{2}, Ludovic Liétard\inst{2},
\\
Boutheina Ben Yaghlane\inst{3}, Hend Ben Hadji\inst{4}}
\authorrunning{S. Jendoubi et al.}

\institute{LARODEC, ISG Tunis, Université de Tunis \and IRISA, Université de
Rennes I \and LARODEC, IHEC Carthage, Université de Carthage
\and
Centre d'Etude et de Recherche des Télécommunications\\
\mailsa\\
\mailsb\\
\mailsc\\}
\maketitle
\begin{abstract}
Social messages classification is a research domain that has attracted
the attention of many researchers in these last years. Indeed, the
social message is different from ordinary text because it has some
special characteristics like its shortness. Then the development of
new approaches for the processing of the social message is now essential
to make  its classification more efficient. In this paper, we are mainly interested in the classification
of social messages based on their spreading on online social networks
(OSN). We proposed a new distance metric based on the Dynamic Time Warping distance and we use it with the
probabilistic and  the evidential $k$ Nearest Neighbors ($k$-NN) classifiers to classify propagation networks (PrNets) of messages.
The propagation network is a directed acyclic graph (DAG)
that is used to record propagation traces of the message, the
traversed links and their types.
We tested the proposed metric with the chosen $k$-NN classifiers on real world propagation traces that
were collected from Twitter social network and we got good classification accuracies.
\end{abstract}
\keywords{Propagation network (PrNet), classification, Dynamic Time Warping (DTW), $k$ Nearest Neighbor ($k$-NN).}

\section{Introduction}

During the past decade, many classification methods have been appeared,
like $k$ Nearest Neighbors ($k$-NN), Naive Bayes, Support Vector
Machines (SVM), etc. Those methods have been applied to several problems
among them text classification and they proved their performance, \cite{Sebastiani2002}. However, when working
with short text like online communications, chat messages, tweets,
etc, we are face to a new challenge. In fact, in a short text there
is no sufficient word occurrences or shared context for a good similarity
measure. Let's take Twitter for example, Twitter is a micro-blogging
service that allows its users to share messages of 140 characters
that are called \textit{tweets}. As a consequence, using a traditional
text classification technique to classify tweets, like the ``Bag-Of-Words''
method, fail to achieve good classification rates due to the message shortness.
Existing works on classification of short text integrate meta-information
from external sources like Wikipedia, World Knowledge
and MEDLINE \cite{Banerjee07,Hu09,Phan09}. They
tend to enrich the content of the message.

The purpose of this paper is to classify social messages without any
access to their content. Our work is motivated by two facts; first, it
is not always possible to have access to the content of the message
but we may have access to its propagation traces, in such
a case, our approaches are useful. Another motivation is that, text
processing techniques, always, need a pre-processing step in which
it is necessary to remove URLs, stop words, questions, special characters,
etc. When working with tweets, for example, after the pre-processing
step, it falls, very often,  on empty messages. Those empty messages
can not be classified by a text based classification technique. Hence
comes the necessity of new classification approaches that consider the propagation of the message. 

Our work is driven by the motivations above, and it achieves the following
contributions: 1) we adapted the Dynamic Time Warping (DTW) distance
\cite{Petitjean12} to be used to measure the distance between
two propagation networks (PrNet for short)%
\footnote{We call propagation network the network that conserves
propagation traces of the message, \textit{i.e.} traversed links and nodes}. 2) we proposed to incorporate the proposed distance
in the probabilistic $k$-NN and the evidential $k$-NN \cite{Denoeux95a} to classify propagation
networks of social messages. Then 3) we tested the classifiers
on real world propagation traces collected from Twitter social network.

This paper is organized as follow: Section 2 discusses some related works. Section 3 provides relevant background. Section 4 introduces the proposed PrNet-DTW distance. And in Section 5 presents results from our experiments. 

\section{Related works}

\subsection{Content based approaches}

Methods that are used for text classification or
clustering always have some limitation with short text, in fact, in
short text there is no sufficient word occurrences. Then, traditional
methods are not suitable for the classification of the social message
that is characterized by its shortness. For example, the use of the
traditional ``Bag-Of-Words'' method to classify tweets may fail to achieve good classification rates. This limitation has attracted
the attention of many researchers who developed several approaches.
The authors in \cite{Sriram10} classified tweets to ``News'', ``Events'',
``Opinions'', ``Deals'' and ``Private Messages'' using a set of
features among them author information and features extracted from
the tweet. In \cite{Banerjee07} and \cite{Hu09}, the authors propose
approaches for short text clustering that use not only the content
of the text but also an additional set of items that is extracted
from an external source of information like Wikipedia and World Knowledge.
Also, \cite{Phan09} classify short and sparse text using a large
scale external data collected from Wikipedia and MEDLINE.

Social messages are, also, classified for sentiment analysis and opinion
mining purposes \cite{Lo09}. The task here, is to identify the dominant
opinion about a product or a brand using text mining techniques. The
author of \cite{Mostafa13} used 3516 tweets to identify costumer's
sentiment about some well known brands. In \cite{He13}, authors used
text published on Twitter and Facebook to analyze the opinion about
three chain of pizza. The reader can refer to \cite{Othman14} for
a recent survey. 

Our work is different from all of the above in that we propose to
classify the social message without access to its content. In fact,
we predict the class of the message by interpreting its propagation
traces through the social network. We think that the proposed approaches
will be useful in the case where there is no access to the content
of the message or when text based methods are unable to classify the
message due to its shortness.

\subsection{Propagation based approaches}

Now we move to present two methods
that were used to classify propagation networks and that were published
in \cite{Jendoubi14a}. The first method uses the probability theory
and the second one incorporates the theory of belief functions. As
we said above, existing classification approaches that are used for
text classification and characterization, always, have some limitation
with short text. To overcome this limitation, we propose to classify
the propagation traces of the message instead of its content. For
an illustrative example, when you receive a letter from your bank,
it is likely to be about your bank account. 

The PrNet classifiers work in two main steps, the first step, is used
to learn the model parameters and the second step, uses the learned
model to classify new coming messages (propagation network of the
message). Both methods have the same principle in the two steps. In
the parameter learning step, we need a set of propagation networks,
PrNetSet that is used to estimate a probability distribution defined
on types of links for each level%
\footnote{We call propagation level the number of links between the source of
the message and the target node.}. 
In the belief PrNet classifier, we use the consonant transformation
algorithm, also called inverse pignistic transformation, \cite{Aregui07a,Aregui08a}
that allows us to transform the probability distribution (output of
the probabilistic parameter learning step) to a BBA distribution while
preserving the least commitment principle \cite{Smets00a}. Once model's
parameters are learned, we can use it to classify a new message (propagation
network of the message). The reader can refer to \cite{Jendoubi14a}
for more details.

These classifiers need a transit step through a compact structure that assigns a probability distribution to each propagation level. This step leads to a loss of information that may be significant in the classification step. Another drawback is that these methods do not work with continuous types of links and a discretization step is always needed in such a case. We think that the proposed PrNet-DTW classifiers will avoid these problems.

\section{Background}

\subsection{Theory of belief functions}

The \textit{Upper and Lower probabilities} \cite{Dempster67a} is the first ancestor of the evidence
theory, also called Dempster-Shafer theory or theory of belief functions.
Then \cite{Shafer76} introduced the \textit{mathematical theory of
evidence} and defined the basic mathematical framework of the evidence
theory, often called \textit{Shafer model}. The main goal of the Dempster-Shafer
theory is to achieve more precise, reliable and coherent information.

Let $\Omega=\left\{ s_{1},s_{2},...,s_{n}\right\} $ be the \textcolor{black}{frame
of discernment.} The basic belief assignment (BBA), $m^{\Omega}$,
represents the agent belief on $\Omega$. $m^{\Omega}\left(A\right)$
is the mass value assigned to $A\subseteq\Omega$, it must respect:
$\sum_{A\subseteq\Omega}m^{\Omega}\left(A\right)=1$. In the case
where we have $m^{\Omega}(A)>0$, $A$ is called focal set of $m^{\Omega}$. 

Combination rules are the main tools that can be used for information
fusion. In fact, in real world applications, we do not have the same
kind of information to be combined, that's why the same combination
rule may performs well in some applications and may gives unsatisfiable
results with other applications. Among these combination rules, we find the Dempster's rule \cite{Dempster67a}, the conjunctive rule of combination (CRC) \cite{Smets90a,Smets93a} and the disjunctive rule of combination (DRC) \cite{Smets93a}.

\subsection{$k$ Nearest Neighbors}

In this paper, we choose the $k$ nearest neighbors classification technique because it is distance based. It will be used to classify propagation traces of social messages together with the proposed distance. In this section we present two $k$-NN based approaches which are the probabilistic $k$-NN and the evidential $k$-NN.

 \textbf{Probabilistic $k$ nearest neighbors}  ($k$-NN)
is a well known supervised method that is generally used for classification.
It needs as input a set of training examples that we know their features
values and their classes, and of course the object to be classified.
Besides we have to specify a measure of distance that will be used
to quantify the matching between the new object $x$ and every object in
the training set. First, the $k$-NN starts by computing the distance between
 $x$ and every object in the training set, then, it selects the $k$ nearest neighbors, {\em i.e.} that have the shortest distance with $x$. Finally, the object $x$ is
classified according to the majority vote principle, {\em i.e.} the algorithm chooses the class that has the maximum occurrence count in the $k$ nearest neighbors set to be the class of $x$. The $k$-NN technique
is surveyed in \cite{Bhatia10}.

\textbf{Evidential $k$ Nearest Neighbors} is an extension
of the probabilistic $k$-NN to the theory of belief functions \cite{Denoeux95a}. The
probabilistic $k$-NN uses distances between the object $x$, to be
classified, and objects in the training set to sort the training example,
then it chooses the $k$ nearest neighbors to $x$. However, according
to \cite{Denoeux95a}, the distance value between $x$ and its nearest
neighbors may be significant. The evidential $k$-NN differs from the probabilistic one in the decision
rule. Let $\Omega=\left\{ s_{1},s_{2},...,s_{n}\right\}$ the set of
all possible classes, be our frame of discernment and $d_{j}$ be
the distance between $x$ and the $j^{th}$ nearest neighbor. The
idea behind the evidential $k$-NN consists on representing each object
of the $k$ neighbors by a BBA distribution defined by: 
\begin{eqnarray}
m\left(\left\{ s_{i}\right\} \right) & = & \alpha\\
m\left(\Omega\right) & = & 1-\alpha\\
m\left(A\right) & = & 0\,\forall A\in2^{C}\setminus\left\{ C_{i}\right\} 
\end{eqnarray}
such that $0<\alpha<1$. If $d_{j}$ is big, $\alpha$ have to be
small. Then it will be calculated as follow: 
\begin{eqnarray}
\alpha & = & \alpha_{0}\Phi_{i}\left(d_{j}\right)\\
\Phi_{i}\left(d_{j}\right) & = & e^{-\gamma_{i}d_{j}^{\beta}}
\end{eqnarray}
where $\gamma_{i}>0$ and $\beta\in\left\{ 1,2,\ldots\right\} $.
After estimating a BBA distribution for each nearest neighbor, the
decision about the class of $x$ is made according to the following steps; first
we combine all BBA distributions using a combination rule. Second,
we apply the pignistic transformation, \cite{Smets05b}, in order
to obtain a pignistic probability distribution. And finally, we choose
the class that have the biggest pignistic probability. In the next
section, we will introduce the dynamic time warping distance and its
extension to compute similarity between propagation networks.

\section{Proposed dynamic time warping distance for propagation networks similarity\label{sec:Proposed-PrNet-DTW}}

The propagation network is a graph based data structure that is used
to store propagation traces of a message. The PrNet has two main characteristics that distinguish it from an ordinary DAG\footnote{Directed Acyclic Graph}; first, its arcs are weighted
by the type of the relationship between users, and second, its paths are time dependent. In this paper, we choose to use distance based classifiers; the probabilistic and the evidential $k$-NN, then, we need to measure the distance between the PrNet to be classified and the training set. In \cite{Jendoubi14a}, we presented two PrNet classifiers that are based on mathematical distances like the Euclidean distance and the Jaccard distance. This solution need to transform the PrNet to a set of probability or BBA distributions, then it computes the distance between those distributions instead of PrNets. This transformation
may lead to a loss of the information. A second solution may be to
use a graph distance metric to measure the similarity between PrNets.
In the literature, we found several distances like \textit{Graph
edit distances} \cite{Gao10}, and \textit{Maximal common sub-graph
based distances} \cite{Bunke02}. However, all these distances do
not consider the time dimension which is a character of the PrNet.
Then comes the need of a new distance that is adapted to weighted
time dependent DAGs like the PrNet. As a solution to this problem
we propose the Dynamic Time Warping distance for propagation networks
similarity (PrNet-DTW).

The Dynamic Time Warping similarity measure \cite{Sakoe71} was first
proposed for speech recognition, it consider the fact that the speech
is time dependent. Recently, \cite{Petitjean12} propose to use it
to measure the similarity between two sequences, \textit{i.e.} a sequence
is an ordered list of elements. DTW distance is used to
consider the order of appearance of each element in the sequences
while computing the distance between them. Let $A=\left(a_{1},a_{2},\ldots,a_{S}\right)$
and $B=\left(b_{1},b_{2},\ldots,b_{T}\right)$ be two sequences. $DTW\left(A_{i},B_{j}\right)$
is the DTW distance between $A$ and $B$ and it is defined as \cite{Petitjean12}:
\vspace{-0.4cm}
\begin{equation}
DTW\left(A_{i},B_{j}\right)=\delta\left(a_{i},b_{j}\right)+\min\begin{cases}
DTW\left(A_{i-1},B_{j-1}\right)\\
DTW\left(A_{i},B_{j-1}\right)\\
DTW\left(A_{i-1},B_{j}\right)
\end{cases}
\end{equation}
\vspace{-0.4cm}

Note that $\delta\left(a_{i},b_{j}\right)$ is a the distance between
the two elements $a_{i}\in A$ and $b_{j}\in B$. As mentioned in \cite{Petitjean12}, the implementation
of this recursive function leads to exponential temporal complexity.
They propose the memoization technique
 as a solution to speed up
the computation. Hence, we need a $\mid S\mid\times\mid T\mid$ matrix
in which we record previous results in order to avoid their computation
in next iterations. This computation technique maintain the time and
space complexity of the DTW distance to $O\left(\mid S\mid\times\mid T\mid\right)$.

The PrNet-DTW distance is used to measure the distance
between two propagation networks. In the first step, we transform
each PrNet to a set of dipaths. We define a dipath as a finite sequence vertices connected with arcs that are directed to the
same direction (line 1 and 2 in algorithm \ref{alg:PN-DTW-algorithm}). 
 We note that all dipaths starts from the source of the message. In the second step,
the PrNet-DTW algorithm loops on the DipathSet1,
at each iteration, it fixes a Dipath and compute
its DTW distance with all Dipaths in DipathSet2 and it takes the minimal
value. Finally, it computes the mean of minimal distances between
Dipaths in DipathSet1 and those in DipathSet2 to be the PrNet-DTW distance. Details are shown
in algorithm \ref{alg:PN-DTW-algorithm}. We choose the $k$-NN algorithm
and evidential $k$-NN algorithm to classify propagation networks
because they are distance based classifiers and they can be used
with the proposed PrNet-DTW distance.

\begin{algorithm*}
\DontPrintSemicolon
\SetKwInOut{Input}{input}\SetKwInOut{Output}{output}
\Input {\textit{PrNet1} and \textit{PrNet2}: Two propagation networks}

\Output {\textit{Distance:} The distance between PrNet1 and
PrNet2.}
\Begin{
\nl$DipathSet1\leftarrow PrNet1.TransformToDipathSet()$\;
\nl$DipathSet2\leftarrow PrNet2.TransformToDipathSet()$\;

\nl\For{$i=1$ \KwTo $DipathSet1.size()$}{

\nl$D\leftarrow maxValue$\;
\nl\For{$j=1$ \KwTo $DipathSet2.size()$}{

\nl$D\leftarrow\min\left(D,\, DTW\left(DipathSet1.get(i),\, DipathSet2.get(j)\right)\right)$\;
\nl$Distance\leftarrow Distance+D$\;
}
\nl$Distance\leftarrow Distance/DipathSet1.Size\left(\right)$; 
}
}
\caption{PrNet-DTW algorithm }

\label{alg:PN-DTW-algorithm} 
\end{algorithm*}

\section{Experiments and results}

We used the library Twitter4j%
\footnote{Twitter4j is a java library for the Twitter API, it is an open-sourced
software and free of charge and it was created by Yusuke Yamamoto.
More details can be found in http://twitter4j.org/en/index.html.%
} which is a java implementation of the Twitter API to collect Twitter
data. We crawled the Twitter network for the period between 08/09/2014
and 03/11/2014. After a data cleaning step, we got our data set that
contains tweets of three different classes: ``Android'',
``Galaxy'' and ``Windows''. To simplify the tweet classification step, we consider a tweet that contains the name of a class \textit{C}, for example a tweet that contains the word ``Android'', of type that class \textit{C}, i.e. the class ``Android'' in our example. Table \ref{tab:Statistics} presents some statistics about the data set.

\vspace{-0.4cm}
\begin{table}
\caption{Statistics of the data set\label{tab:Statistics}}

\begin{centering}
\begin{tabular}{|c|c|c|c|c|c|c|c|}
\cline{2-8} 
\multicolumn{1}{c|}{} & \textbf{\#User} & \textbf{\#Follow} & \textbf{\#Tweet} & \textbf{\#Retweet} & \textbf{\#Mention} & \textbf{\#Prop. links} & \textbf{\#PrNet}\tabularnewline
\hline 
\textbf{Android} & 6435 & 9059 & 81840 & 3606 & 6092 & 7623 & 224 \tabularnewline
\hline 
\textbf{Galaxy} & 4343 & 4482 & 8067 & 2873 & 5965 & 6819 & 161\tabularnewline
\hline 
\textbf{Windows} & 5775 & 12466 & 11163 & 2632 & 3441 & 11400 & 219\tabularnewline
\hline 
\end{tabular}
\par\end{centering}

\end{table}
\vspace{-0.4cm}

The remainder of this section is organized as follow: we present
our experiments configuration, the method with which we extracted
propagation and the computation process of link weights. Then,
we compare the proposed classifiers with those of \cite{Jendoubi14a}.

\subsection{Experiments configuration}

In our experiments, we need to extract propagation traces of each
type of message. Here, we consider that a tweet of type $a$ was propagated
from a user $u$ to a user $v$ if and only if $u$ posts a tweet of type $a$
before $v$ and at least one of these relations between $u$ and $v$
exists: 1) $v$ follows $u$, 2) $u$ mentions $v$ in a tweet of type
$a$, 3) $v$ retweets a tweet of type $a$ written by $u$. After
getting propagation traces we extract propagation networks such that
each PrNet has to have one source.

We define types of links that are used to measure the similarity
between propagation networks. In Twitter social network there are
three possible relations the first one is explicit which is the follow
relation, the second and the third relations are implicit which are
the mention and the retweet. Another property of Twitter, is that
between two users $u$ and $v$ we can have a follow, a mention and/or
a retweet relation. We assign to each of those a weight \cite{BenJabeur13}
and we assign to each link a vector of weights that has the form $\left(w_{f},w_{m},w_{r}\right)$.
Let $S_{u}$ be the set of successor of $u$, $P_{u}$ the set of
predecessor of $u$, $T_{u}$ the set of tweets of $u$, $R_{u}\left(v\right)$
the set of tweets of $u$ that were retweeted by $v$, $M_{u}\left(v\right)$
the set of tweets of $u$ in which $v$ was mentioned and $M_{u}$
the set of tweets in which $u$ mentions another user. We compute weights
\cite{BenJabeur13} as follow:
\begin{itemize}
\item Follow relation: $w_{f}\left(u,v\right)=\frac{\mid S_{u}\cap\left(P_{u}\cap\left\{ u\right\} \right)\mid}{\mid S_{u}\mid}$
\item Mention relation: $w_{m}\left(u,v\right)=\frac{\mid M_{u}\left(v\right)\mid}{\mid M_{u}\mid}$
\item Retweet relation: $w_{r}\left(u,v\right)=\frac{\mid R_{u}\left(v\right)\mid}{\mid T_{u}\mid}$
\end{itemize}
Finally, we choose the euclidean distance to evaluate the $\delta\left(a_{i},b_{j}\right)$ in the computation process of the PrNet-DTW.

\subsection{Experiments evaluation}

In our experiments, we want to evaluate the performance of the PrNet-DTW
distance, then, we integrate it in the $k$-NN and
the evidential $k$-NN classifiers and we compare the proposed classifiers
with those proposed in \cite{Jendoubi14a}. As PrNet classifiers works
with a discrete types of links \cite{Jendoubi14a}, a discretization
step was needed, \textit{i.e.} if the weight value ($w_{f},w_{m}$ or $w_{r}$) is greater than 0 we replace it by 1 in the discrete weight vector elsewere we replace it by 0.
For example, if the link is weighted by the vector  
$\left(w_{f}=0.5,\, w_{m}=0,\, w_{r}=0.25\right)$, the output after the discretization step will be $\left(1, \, 0, \, 1\right)$.
 In the remainder of our experiments, we divide, randomly, our data set
into two subsets; the first one contains 90\% of PrNets and it is used
for training and the second one (10\%) is used for testing.

\begin{figure}[t]
\begin{centering}
\includegraphics[scale=0.65]{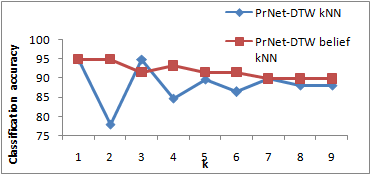}
\par\end{centering}

\caption{$k$ variation\label{fig:k-variation}}

\end{figure}

The algorithm $k$-NN is known to be dependent to $k$ value, and
varying $k$ may vary the classification accuracy. Then, to see the
impact of the parameter $k$, we made this experiment; we run our
$k$-NN based algorithms with multiple $k$ values and we obtained results
in Figure \ref{fig:k-variation}. We note that odd values are more
appropriate to $k$ when we use PrNet-DTW Probabilistic $k$-NN. Moreover,
the PrNet-DTW belief $k$-NN has not the same behavior as the PN-DTW
Probabilistic $k$-NN. In fact, the curve of the evidential classifier
is more stable than the curve of the probabilistic one and the variation
of the value of $k$ does not have a great effect on the classification
accuracy.

A second experiment was done to evaluate and compare the proposed classification
methods. We fixed the parameter $k$ to 5 and we obtained results
in table \ref{tab:Comparison}. As shown in table \ref{tab:Comparison},
the probabilistic and the belief classifiers do not give good classification
accuracy, this behavior is a consequence of the discretization step
that leads to the loss of the information given by weights values.
In contrast, the PrNet-DTW based classifiers show their performance,
indeed, we have got good accuracy rates: 88.69\% ($\pm3.39$, for a 95\% confidence
interval) and 89.92\% ($\pm3.20$) respectively. We see also that the
PrNet-DTW belief classifier gives slightly better results.

\begin{table}[t]
\caption{Comparison between PrNet classifiers\label{tab:Comparison}}

\centering{}%
\begin{tabular}{|>{\centering}p{25mm}|>{\centering}p{22mm}|>{\centering}p{22mm}|>{\centering}p{22mm}|>{\centering}p{22mm}|}
\cline{2-5} 
\multicolumn{1}{>{\centering}p{25mm}|}{} & \textbf{Proba classifier} & \textbf{Belief classifier} & \textbf{PrNet-DTW $k$-NN} & \textbf{PrNet-DTW Belief $k$-NN}\tabularnewline
\hline 
\textbf{Accuracy} & {51.97\% $\pm$2.04} & {52.25\% $\pm1.99$} & \textbf{88.69\% $\pm3.39$} & \textbf{89.92\% $\pm3.20$}\tabularnewline
\hline 
\end{tabular}
\end{table}
\vspace{-0.4cm}

\section{Conclusion}

To sum up, we presented a new distance metric that we called
PrNet-DTW. Our measure is used to quantify the distance between propagation
networks. Also, we showed the performance of our measure in the process
of classification of propagation networks, indeed, we defined two
classification approaches that uses the PrNet-DTW measure which are
the probabilistic $k$-NN and the evidential $k$-NN.

For future works, we will search to improve the PrNet-DTW based classifiers
by taking into account the content of the message to be classified,
in fact, we believe that a classification approach that uses information
about the content of the message and information about its propagation
will further improve the results.

\section{Acknowledgement}
These research works and innovation are carried out within the framework
of the device \textit{MOBIDOC} financed by the European Union under the \textit{PASRI}
program and administrated by the \textit{ANPR}. Also, we thank the "\textit{Centre d'Etude et de Recherche des Télécommunications}" (CERT) for their support.%

\bibliographystyle{splncs03}
\bibliography{biblio}

\end{document}